\documentclass[letterpaper, 10 pt, conference]{ieeeconf}  
\IEEEoverridecommandlockouts                              

\usepackage{graphics} 
\usepackage{epsfig} 
\usepackage{times} 
\usepackage{amsmath} 
\usepackage{amssymb}  
\usepackage{bm}
\usepackage{multirow}
\usepackage{float}
\usepackage{subfig}
\usepackage{url}
\usepackage{hyperref}
\hypersetup{hypertex=true,
hidelinks=true,
urlcolor=blue,
colorlinks=true,
linkcolor=blue,
anchorcolor=blue,
citecolor=blue}

\title{\LARGE \bf
Robot Trajectron: Trajectory Prediction-based Shared Control for Robot Manipulation
}

\author{Pinhao Song$^{1}$, Pengteng Li$^{3}$, Erwin Aertbeliën$^{1,2}$, Renaud Detry$^{1}$
\thanks{$^{1}$P. Song, E. Aertbeliën, and R. Detry are with Robotics Research Group, the Department of Mechanical Engineering, KU Leuven, Belgium.
        {\tt\small Email: firstname.lastname@kuleuven.be}}
\thanks{$^{2}$Flanders Make@KU Leuven    }    
\thanks{$^{3}$P. Li is with College of Computer Science and Software Engineering, Shenzhen University, Shenzhen, China.
        {\tt\small Email: 2110276192@email.szu.edu.cn}}
}

\begin{document}

\maketitle
\thispagestyle{empty}
\pagestyle{empty}

\begin{abstract}
We address the problem of (a) predicting the trajectory of an arm reaching motion, based on a few seconds of the motion's onset, and (b) leveraging this predictor to facilitate shared-control manipulation tasks, easing the cognitive load of the operator by assisting them in their anticipated direction of motion. Our novel intent estimator, dubbed the \emph{Robot Trajectron} (RT), produces a probabilistic representation of the robot's anticipated trajectory based on its recent position, velocity and acceleration history. Taking arm dynamics into account allows RT to capture the operator's intent better than other SOTA models that only use the arm's position, making it particularly well-suited to assist in tasks where the operator's intent is susceptible to change. We derive a novel shared-control solution that combines RT's predictive capacity to a representation of the locations of potential reaching targets. Our experiments demonstrate RT's effectiveness in both intent estimation and shared-control tasks. We will make the code and data supporting our experiments publicly available at \url{https://github.com/mousecpn/Robot-Trajectron.git}.
\end{abstract}

\section{INTRODUCTION}
As robotic hardware improves, teleoperated robot applications emerge at an increasing rate in domains as varied as subsea maintenance, surgery, or assistive devices. While simple, direct teleoperation is at times feasible, integrators often prefer a form of shared control, where a human operator and an autonomous agent work in tandem, reducing the cognitive load of the operator, and/or improving safety or performance by filtering operator noise and exploiting sensor feedback at a rate that surpasses human capacity. For instance, in Brain-Computer-interface (BCI) controlled robot manipulation, the inherent noise in brain signals leads to considerable effort on the part of the patient to realize precise manipulation. With shared control, the user can achieve their goal with increased smoothness and eased effort.

Anticipating the user's intended motion during execution is a crucial component of the shared control paradigm. This ability is usually referred to as intent estimation.
Current intent estimators assume that the user has a predefined goal and maintains a consistent intent while taking actions to achieve that goal, which does not always hold true. Furthermore, most intent estimators rely on position-based methods, which consider only the distances between past positions (or the next intended position) and each goal to infer the user's intent \cite{handmove, xu2020shared}, ignoring the robot dynamics. For example, position-based MaxEnt IOC like \cite{aarno2008motion, muelling2017autonomy, gottardi2022shared} assumes that the user approximately optimizes a cost function which is the cumulative distance between the robot and the goal, and infers the user's intent based on this assumption.

This paper addresses the aforementioned challenges through two primary contributions:
First, we propose \emph{Robot Trajectron} (RT), a model that anticipates the trajectory of a robot's end-effector during a reach-to-grasp motion, i.e., RT predicts the end-effector's \emph{future trajectory}. RT bases its prediction on the motion's recent dynamics (position, velocity, and acceleration in the past few seconds), by contrast to prior works that only consider the positions of waypoints of the arm's recent motion \cite{handmove, xu2020shared}. Our model is data-driven, and it learns the robot behavior with few strict assumptions. This characteristic allows it to make predictions with short-term historical dynamics while maintaining noise resiliency, resulting in a fast response to a change of intent, which contrasts with prior work that assumes a fixed intent \cite{aarno2008motion, muelling2017autonomy, gottardi2022shared}.

Second, we also propose a novel shared-control paradigm that leverages RT as an intent estimator. Our shared-control paradigm follows the basic idea of Artificial Potential Fields (APFs) \cite{apf} to guide the robot towards its goals. To flexibly adjust the authority of the user, we propose a straightforward agreement mechanism that reinforces the RT's authority in cases of consensus, yet permits the user to override in case of conflict.
To assess the efficacy of the approach, comprehensive experiments are conducted in both simulated data and real-world teleoperation tasks. We show that the proposed shared-control paradigm outperforms the prevalent MaxEnt IOC, especially in the case of intent change.

To summarize, the main contributions of this paper are:
\begin{itemize}
    \item A trajectory prediction model Robot Trajectron, which considers the dynamics of the robot's motion and outputs a probabilistic representation of its future motion. In addition, in the specific case of a tabletop scenario, we provide a means of mapping RT's prediction onto the objects that stand on the table.
    \item A novel shared-control paradigm is proposed to assist the operator in approaching a goal (an object) that lies near the predicted trajectory.
    \item Comprehensive experimental validations are conducted in both a simulation and a real-world grasping task to show the effectiveness of our method.
\end{itemize}
 
\section{RELATED WORK}
\noindent \textbf{Trajectory Prediction.}
Trajectory prediction involves estimating future trajectories based on observed paths, which applies not only to pedestrians \cite{nsp,socialattn,wang2007gaussian} but also to vehicles \cite{katuwandeniya2022exact,trajectron++}. 
Recently, deep learning methods, due to their strong ability to model social interactions and agents' momentum, have largely outperformed traditional methods.
Recurrent Neural Networks (RNNs) have been explored first due to their ability to process sequential data \cite{sociallstm,bartoli2018context}.
However, given past trajectories, there can be numerous potential future trajectories. Early RNN-based methods like Social-LSTM \cite{sociallstm} which can only generate a single path, fail to capture the multimodal nature of trajectory prediction, which limits its applicability.
To handle this challenge, generative architectures have been introduced into trajectory prediction, including Generative Adversarial Network (GAN) \cite{socialgan} and Conditional Variational Autoencoder (CVAE) \cite{mangalam2020not,trajectron,trajectron++}. For instance, Trajectron \cite{trajectron} follows the CVAE framework and models future velocities using Gaussian Mixture Models (GMMs). This approach provides an explicit distribution of all possible trajectories, offering practicality and flexibility in various applications.

Trajectron shows promise as a robot manipulation intent estimator, but it faces various challenges. The first challenge stems from dealing with 3D data instead of 2D scenarios for which it was designed. Another challenge lies in utilizing Trajectron's output to understand the intent. To overcome these difficulties, RT works in 3D space, and maps the predicted trajectory to the distribution of potential goals (i.e., objects).


\noindent \textbf{Intent Estimation in Shared Control.}
To efficiently assist users, it is crucial for the system to comprehend their intent. Early studies \cite{vanhooydonck2003shared,goodrich2003seven} suggest that mandating explicit intent specification is inefficient and sometimes unfeasible (e.g., BCI-controlled setting). Consequently, contemporary research places emphasis on harnessing implicit cues such as user commands and environmental sensing to deduce user intent.
One prevalent approach is to employ a Hidden Markov Model (HMM) for intent inference, treating intent as the model's latent state \cite{ding2011human,aarno2005adaptive,aarno2008motion}. Additionally, Bayesian networks \cite{schrempf2007tractable,tahboub2006intelligent} have also been explored for intent estimation. 
One of the intent estimation milestones is MaxEnt IOC \cite{ziebart2008maximum}, which inspires a lot of shared control works and achieves promising performance in the cluttered environment \cite{javdani2018shared,muelling2017autonomy,gottardi2022shared}. MaxEnt IOC assumes that the user is an intent-driven agent who seeks to optimize a cost function which is the exponential of the reward. The prevailing way to design to reward is to use the negative distance between the robot and the goal \cite{aarno2008motion}. The distribution over goals can be inferred from the likelihood which is mapped from the rewards of all previous steps. 

However, the methods mentioned above ignore motion dynamics and follow a consistent-intent assumption that does not always stand. In this paper, we build a shared control system based on RT, which provides assistance in reaching the goal along a predicted trajectory and promptly adapts to the intent change by considering dynamic information.

\section{ROBOT TRAJECTRON}
This section introduces our intent-prediction model \emph{Robot Trajectron} (RT). We consider a scenario where an operator wishes to move a manipulator towards one of multiple objects sitting on a table. RT assumes that the user is in the act of guiding the robot from a starting position (usually a rest position) towards a goal (one of the objects). The model predicts the robot's expected future trajectory based on the trajectory it has followed from its start position to its current position. In addition, the model also produces a map that shows where the trajectory is likely to intersect with the table plane.

\subsection{Model Architecture}

RT models the probability distribution of the robot's future trajectory, conditioned on the trajectory it has followed to this point. Let us denote the position, velocity, and acceleration of the gripper at a time $t$ with $X^{t}$, $\dot{X}^{t}$, and $\ddot{X}^{t}$. Let us also denote by $\bm{x} = [X^{(1:T_\textnormal{obs})}, \dot{X}^{(1:T_\textnormal{obs})}, \ddot{X}^{(1:T_\textnormal{obs})}] \in \mathbb{R}^{T_\textnormal{obs} \times 9}$ the history of position, velocity and acceleration from the beginning of the motion to the current time, and by $\bm{y} = \dot{X}^{(T_\textnormal{obs}+1:T_\textnormal{obs}+T)} \in \mathbb{R}^{T \times 3}$ the future velocity.

Our aim is to model $p(\bm{y}|\bm{x})$, i.e., future velocities conditioned on past positions, velocities and accelerations.
As noted in the literature \cite{cvae}, a simple RNN representation of $p(\bm{y}|\bm{x})$ may struggle with multimodal distribution, i.e., cases where multiple future trajectories are compatible with a single past trajectory.
Instead, we mimic the CVAE framework \cite{cvae, trajectron} and introduce a latent variable $\bm{r}$, to facilitate the encoding of a low-dimensional, multi-modal representation of trajectory data:
\begin{equation}
    p(\bm{y}|\bm{x})=\sum_{\bm{r}} p_{\psi}(\bm{y}|\bm{x},\bm{r}) p_{\theta}(\bm{r}|\bm{x}). \label{marginalization}
\end{equation}
We encode the probability distributions shown above with neural networks, and tune their parameters to maximize the likelihood of a dataset $(\bm x^{(i)}, \bm y^{(i)})$ by maximizing, per CVAE practice \cite{beta-vae,cvae}, the $\beta$-weighted evidence-based lower bound (ELBO):
\begin{equation}
\begin{aligned}
    \mathop{\textnormal{max}}_{{\theta}, {\psi},{\phi}} & ~\mathbb{E}_{\bm{r}\sim q_{\phi}(\bm{r}|\bm{x},\bm{y})}[\mathop{\textnormal{log}}p_{\psi}(\bm{y}|\bm{x},\bm{r})] \\
    &- \beta D_{KL}(q_{\phi}(\bm{r}|\bm{x},\bm{y})||p_{\theta}(\bm{r}|\bm{x})), \label{eq:elbo}
\end{aligned}
\end{equation}
where $q_{\phi}(\bm{r}|\bm{x},\bm{y})$ approximates $p_{\theta}(\bm{r}|\bm{x})$, and ${\theta}$, ${\phi}$ and ${\psi}$ denote the learnable parameters of the neural representation underlying $p_{\theta}$, $q_{\phi}$ and $p_{\psi}$.

\begin{figure}[t]
  \centering
  \includegraphics[width=\linewidth]{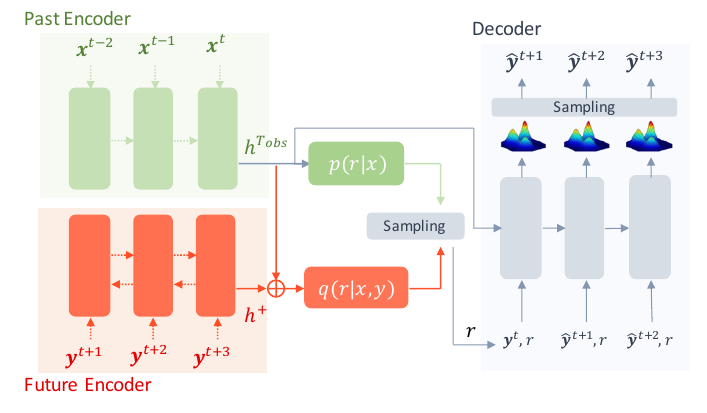}
  \caption{The architecture of the Robot Trajectron. 
  The red lines denote the train-only operations, while the green lines denote the predict-only operations. See text for details.
  }
  \label{rt}
\end{figure}
In accordance with the CVAE framework, $p_{\theta}$, $q_{\phi}$ and $p_{\psi}$ are probability distributions. We model $p_{\theta}$ and $q_{\phi}$ with Bernouilli distributions whose parameters are generated with multi-layer perceptrons (MLPs) fed by LSTM trajectory encoders (see Fig.\ \ref{rt}). We denote by $\theta_{{\ell}}$ and $\phi_{{\ell}}$ the parameters of the two LSTMs that encode the past and future trajectories respectively, and by $\theta_{\textnormal{m}}$ and $\phi_{\textnormal{m}}$ the parameters of the two corresponding MLPs, with $\theta = (\theta_{{\ell}}, \theta_{\textnormal{m}})$ and $\phi = (\phi_{{\ell}}, \phi_{\textnormal{m}})$. Formally, the Bernoulli parameters of $p_{\theta}$, denoted with $B_\theta$, are obtained with:
\begin{equation}
    B_\theta = \textnormal{MLP}(\bm{h}^{T_\textnormal{obs}};{\theta_{\textnormal{m}}}),
\end{equation}
where $\bm{h}^{T_\textnormal{obs}}$ is derived by the past-trajectory LSTM as
\begin{equation}
    \bm{h}^{t} = \textnormal{LSTM}(\bm{h}^{t-1}, \bm{x}^{t};\theta_{{\ell}}).\label{eq:lstmpast}
\end{equation}
The Bernouilli parametres of $q_{\phi}$, denoted with $B_\phi$, are obtained with
$
    B_\phi = \textnormal{MLP}([\bm{h}^{T_\textnormal{obs}};\bm{h}^{+}];{\phi_{\textnormal{m}}}),
$
 where $\bm{h}^{T_\textnormal{obs}}$ is obtained with Eq.\ \ref{eq:lstmpast} and $\bm{h}^{+}$ is obtained with the future-trajectory LSTM as
$
    \bm{h}^{t} = \textnormal{BiLSTM}(\bm{h}^{t-1}, \bm{y}^{t};\phi_{{\ell}}).
$

We model future velocities $p_{\psi}$ with velocity-space Gaussian Mixture Models (GMMs) updated at each timestep. We denote the parameters of the GMMs at time $t$ with $G^t = \{(\bm{\mu}^t_{c}, \bm{\Sigma}^t_{c}, \alpha^t_{c})\}_{c=1}^C$, where $C$ is the number of Gaussian components.
The decoder models future velocities with GMMs parametrized as follows:
\begin{equation}
    [G^t, \bm{h}^{t}] = LSTM([\hat{\bm{y}}^{t-1},\bm{r},\bm{h}^{t-1}];{\psi}), \label{eq:velocitygmm}
\end{equation}
where 
\begin{equation}
    \bm{r} \sim\begin{cases}
    &q_{\phi}(\bm{r}|\bm{x},\bm{y}),\ \text{for training} \\
    &p_{\theta}(\bm{r}|\bm{x}),\ \ \ \ \text{for testing},
     \end{cases} 
\end{equation}
and initializing the decoder with $\bm{h}^{T_{obs}}$.
We predict the velocity at time $t$ via sampling, as $\hat{\bm{y}}^{t} \sim \textnormal{GMMs}(G^t)$. 
We note that instead of encoding $\bm{\Sigma}$ with the six parameters of its matrix representation, we encode it with the six parameters of the lower-triangular matrix $\bm{L}$ of its Cholesky decomposition $\bm{\Sigma} = \bm{L}^T \bm{L}$. 
This representation improves training stability and it allows us to effectively sample from a Gaussian with the simple expression  $\bm{\mu}+\bm{L}\bm{z}$,  $\bm{z}\sim \mathcal{N}(0,1)$.

\subsection{Trajectory Prediction}
Once trained, RT allows us to obtain the most likely future velocities by sampling from its GMMs, as:
\begin{align}
    &\bm{r}_{\text{best}} = \mathop{\textnormal{argmax}}_{\bm{r}}p_{\theta}(\bm{r}|\bm{x}),\\
   & \hat{\bm{y}}_{\textnormal{ml}} = \mathop{\textnormal{argmax}}_{\bm{y}}p_{\psi}(\bm{y}|\bm{x},\bm{r}_{\text{best}}).
\end{align} 
The position trajectory is obtained by integration as:
\begin{equation}
    X_{\textnormal{ml}}^t = X^{t-1}_{\textnormal{ml}} + \dot{X}_{\textnormal{ml}}^{t} \cdot \Delta t,  \label{eq:pos-trajectory}
\end{equation}
where $\hat{\bm{y}}_{\textnormal{ml}} = \dot{X}_{\textnormal{ml}}^{(T_\textnormal{obs}+1:T_\textnormal{obs}+T)}$ and $\Delta t$ is the time interval.

\subsection{Target Selection}

The trajectory derived above \eqref{eq:pos-trajectory} is conditioned on the robot's motion alone. Motion is however not the only cue that informs intent. An understanding of the robot's environment, obtained through vision for instance, often plays a complementary role. In this section, we assume a tabletop scenario and the availability of the locations of target objects disposed on the table, and we discuss means of combining those object data to trajectory data.

We first use RT to compute a probabilistic representation of locations where the robot's motion is likely to intersect with the table plane. To this end, we first convert the velocity GMMs \eqref{eq:velocitygmm} into position GMMs with:
\begin{equation}
\begin{aligned}
   &\bm{\mu}^{t}_{c,p} = \bm{\mu}^{t-1}_{c,p} + \bm{\mu}^{t}_{c} \Delta t,\\ 
   &\bm{\Sigma}^t_{c,p} = \bm{\Sigma}^{t-1}_{c,p} + \bm{\Sigma}^{t}_{c}(\Delta t)^2, \\
   & \alpha^t_{c,p} = \alpha^t_{c},
   \label{update}
\end{aligned}
\end{equation}
where $\bm{\mu}^{t}_{c,p}$, $\bm{\Sigma}^t_{c,p}$, $\alpha^t_{c,p}$ are the mean, covariance and prior of component $c$ at time $t$ in the position GMMs, which are initialized to the last position in the past trajectory and zero matrices at time $T_{\textnormal{obs}}$, respectively.
To derive which object is intended from the position GMMs in 3D space, we take an intersecting section at the table plane $h_{\textnormal{tab}}$ and obtain 2D position GMMs. Given a position of an object denoted as $\bm{g}$, its probability can be obtained from the 2D position GMMs as:
\begin{equation}
p(\bm{g}|\bm{\xi}, h_{\textnormal{tab}}) = \mathop{\textnormal{2DGMMs}}(\bm{g}|\{G^t\}_{t=T_\textnormal{obs}+1}^{T_\textnormal{end}}, h_{\textnormal{tab}}) \label{infer}
\end{equation}
where $\bm{\xi}$ is the past trajectory, and $T_{end}$ is the end time of the process represented by Eq. (\ref{update}). The goal with the highest probability will be the intended goal.

\begin{figure}[t]
  \centering
  \includegraphics[width=\linewidth]{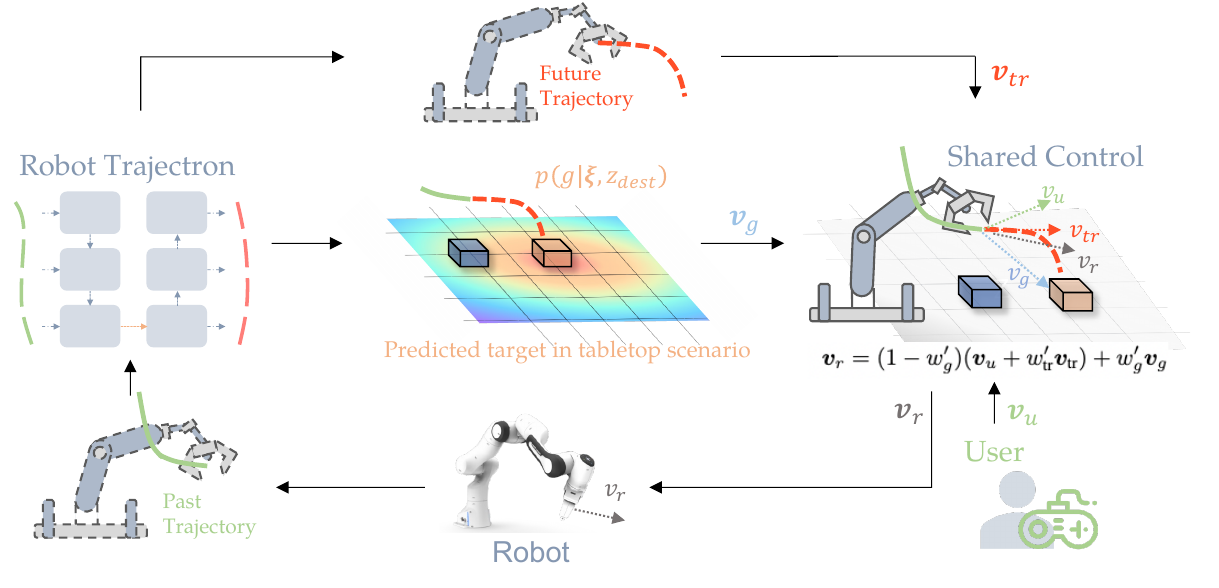}
  \caption{Overview of the proposed shared control. $\bm{v}_g$ denotes the velocity towards the most likely object, given the predicted trajectory. $\bm{v}_{\textnormal{tr}}$ denotes the velocity along the trajectory predicted by RT. $\bm{v}_u$ denotes noisy user command. See text for details.}
  \label{pipeline}
\end{figure}

\section{Shared Control}
In this section, we explain how we use RT to assist the user in a reaching motion. Our method is illustrated in Fig. \ref{pipeline}. The trained RT utilizes the robot's motion as input and produces (a) the most likely trajectory and  (b) a table-plane GMM representation of likely target points. Considering both outputs from RT and the user's command, the shared-control system generates a final velocity command to control the robot.


We design our method based on Artificial Potential Fields (APFs) \cite{apf}, a widely adopted shared-control algorithm, which creates attractive/repulsive fields that guide the motion towards a goal and steer away from obstacles. 
Our solution uses two attractor fields.
First, a Goal Attraction Field (GAF) guides the motion towards the object identified via Eq.\ (\ref{infer}):
\begin{equation}
    U^g_a(\bm{p})=w_g{\parallel \bm{g}-\bm{p} \parallel},
\end{equation}
where $\bm{g}$ is the position of the goal with the highest probability $p(\bm{g}|\bm{\xi},h_{\text{tab}})$, and $\bm{p}$ is the position of the robot. $w_{g}$ is the goal attraction weight, which we calculate as:
\begin{equation}
    w_g = \text{min}(\gamma \cdot p(\bm{g}|\bm{\xi},h_{\text{tab}}), \nu),\label{attraction_goal}
\end{equation}
where $\gamma$ is an amplification coefficient, and $\nu$ is a threshold that we set to $0.1$ in our experiments. 
According to Eq. (\ref{attraction_goal}), the goal attraction weight is linked to the probability generated by RT, which serves as a valuable indicator of uncertainty. If RT is confident that the current motion points unambiguously towards a certain goal, it will strongly push towards that goal.

Our second attractor field helps the robot along RT's predicted trajectory during segments of the motion where Eq.\ (\ref{infer}) does not allow us to confidently select a goal. Assuming that $\bm{p}_{\textnormal{tr}}$ is the first point of the most likely predicted trajectory, we build the Trajectory Following Field (TFF) as:
\begin{equation}
    U^{\textnormal{tr}}_a(\bm{p})=w_{\textnormal{tr}}{\parallel \bm{p}_{\textnormal{tr}}-\bm{p} \parallel},
\end{equation}
\vspace{-0.5cm}
\begin{equation}
    w_{\textnormal{tr}}= \text{max}(\frac{l_{pred}}{l_{past}+l_{pred}}, \zeta),
\end{equation}
where $l_{past}$ and $l_{pred}$ are the length of the past trajectory and the predicted trajectory, respectively, and $\zeta$ is a threshold that we set to $0.7$ in our experiments. $w_{\textnormal{tr}}$ is the trajectory-following weight. TFF can provide more assistance early on in the reaching motion, reducing the noise and stabilizing the prediction of RT.
Combining two fields, we define the robot velocity as:
\begin{equation}
\begin{aligned}
    \bm{v}_{r} & = \bm{v}_{u} - (\nabla U^g_a(\bm{p})+\nabla U^{\text{tr}}_a(\bm{p})) \\
    & = \bm{v}_{u} + w_g\frac{\bm{g}-\bm{p}}{\parallel \bm{g}-\bm{p} \parallel}+w_{\text{tr}}\frac{\bm{p}_{\text{tr}}-\bm{p}}{\parallel \bm{p}_{\text{tr}}-\bm{p} \parallel} \\
    & = \bm{v}_{u} + w_g \bm{v}_{g} + w_{\text{tr}} \bm{v}_{\text{tr}}, \label{shared_control}
\end{aligned}
\end{equation}
where $\bm{v}_{g}$ and $\bm{v}_{\text{tr}}$ are the velocities generated by the GAF and TFF, respectively. $\bm{v}_{u}$ is the user velocity command, and  $\bm{v}_{r}$  is the velocity command sent to the robot. $\bm{v}_{r}$ considers both the position of the intended goal and the predicted trajectory. 

Even though $\bm{v}_{r}$ takes RT's uncertainty into account to trade between user commands and AI assistance, the user will still feel a strong impedance if they change their intent (pick a different goal) when the robot is near one of the goals. To address this issue, we propose an agreement mechanism that balances the weight of the user and AI:
\begin{align}
& a_{g} = \text{max}(\frac{\bm{v}_{u} \bm{v}_{g}}{\parallel \bm{v}_{u}\parallel \parallel \bm{v}_{g}\parallel},0), & w_g' = \sqrt{a_{g} w_g},\\
& a_{\text{tr}} = \text{max}(\frac{\bm{v}_{u} \bm{v}_{\text{tr}}}{\parallel \bm{v}_{u}\parallel \parallel \bm{v}_{\text{tr}} \parallel},0),  & w_{\text{tr}}' = \sqrt{a_{\text{tr}} w_{\text{tr}}}, 
\end{align}
where $a_{g}$ and $a_{\text{tr}}$ are the agreement of goal control and trajectory control. 
The agreement mechanism allows the user to regain authority despite a high RT confidence.
Finally, the user command, the GAF and TFF may at times conflict with one another, causing oscillations. We therefore introduce a soft switch control in Eq. (\ref{shared_control}) as:
\begin{equation}
    \bm{v}_{r} =  (1-w_g')(\bm{v}_{u} + w_{\text{tr}}' \bm{v}_{\text{tr}}) + w_g' \bm{v}_{g}. \label{shared_control2}
\end{equation}
Accordingly, when the confidence of the goal is low, the robot will tend to follow the user command and the predicted trajectory. When the confidence of the goal is high, the robot will tend to be attracted by the intended goal. With the agreement mechanism, the robot will mainly follow the user's command when it conflicts with the AI.




\section{Experiments}
To comprehensively evaluate the proposed method, we conducted one evaluation of RT in simulation, one shared-autonomy experiment on a real robot, and a change-of-intent experiment.
Our research platform for the experiments is Franka Research 3 with a Microsoft Xbox joystick as the control interface.

\begin{figure*}[t]
  \centering
  \includegraphics[width=0.9\linewidth]{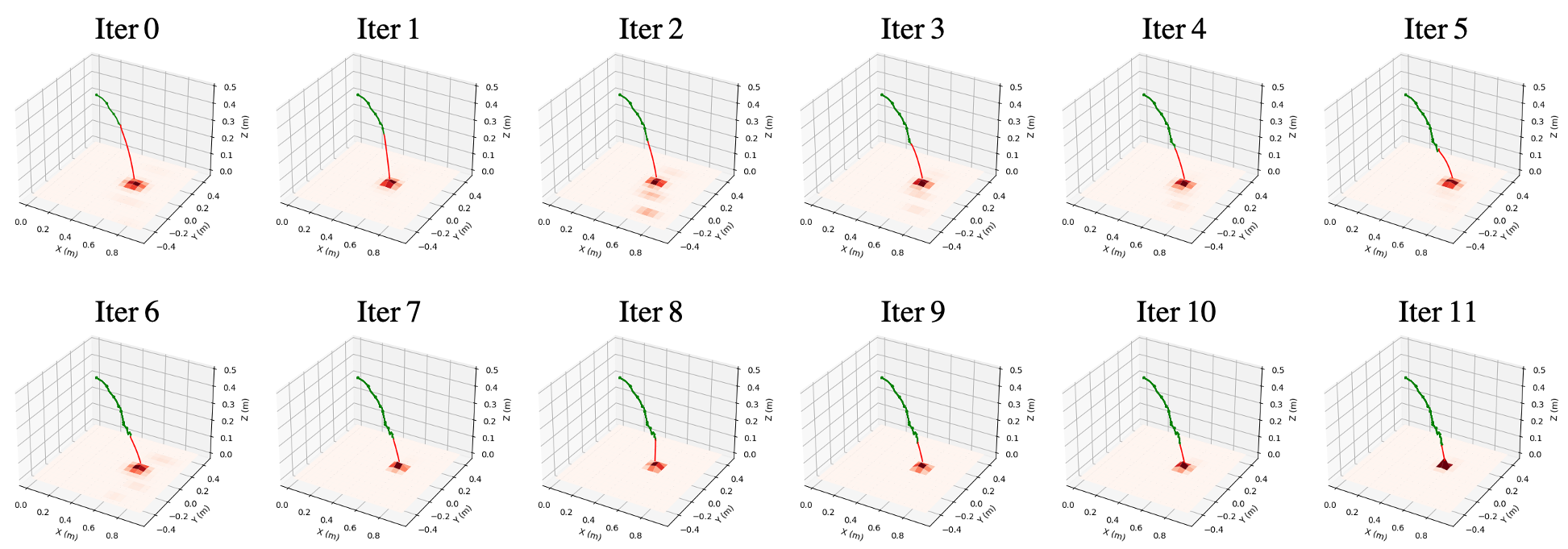}
  \caption{Visualization of RT. The most likely trajectory and the 2D table GMMs are shown. The green line denotes the past trajectory, while the red line denotes the predicted trajectory.}
  \vspace{-0.4cm}
  \label{fig:viz_traj}
\end{figure*}

\begin{figure} [t!]
\centering
    \resizebox{\columnwidth}{!}{
    \subfloat[\label{fig:setup}]{
    \includegraphics[scale=0.3]{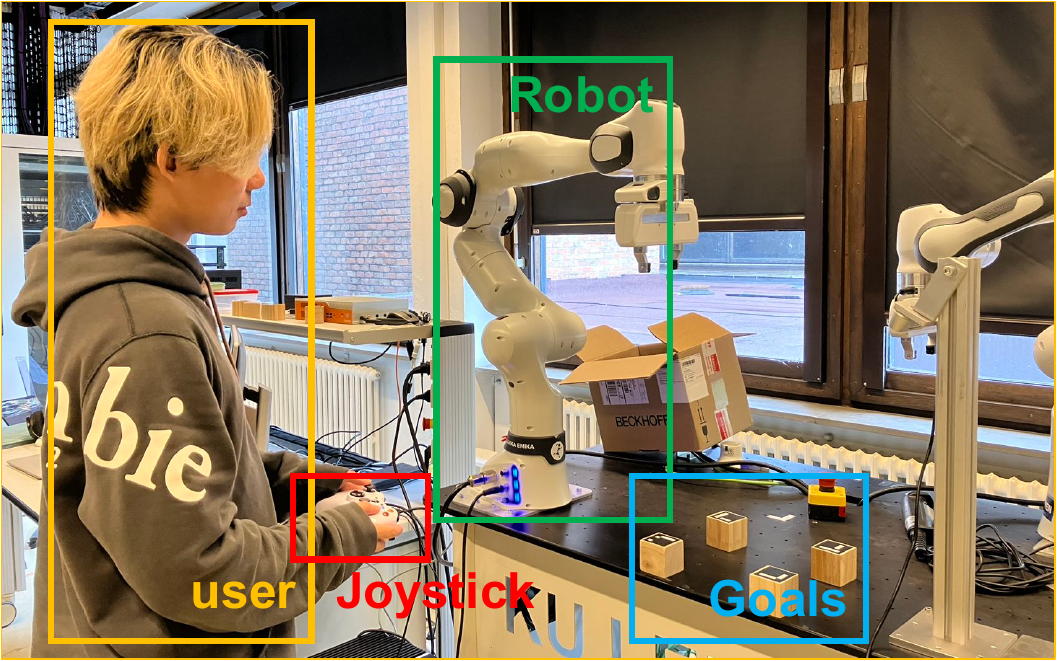}}
    \subfloat[\label{fig:traj}]{
    \includegraphics[scale=0.34]{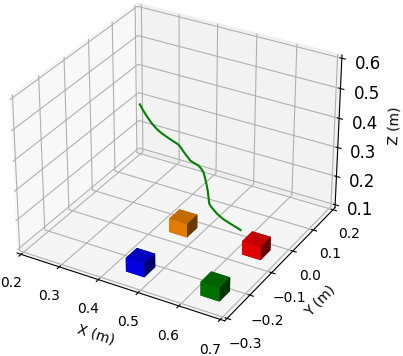}} 
    }
    \vspace{-0.3cm}
    \\
    \resizebox{\columnwidth}{!}{
    \subfloat[\label{fig:weight}]{
    \includegraphics[scale=0.3]{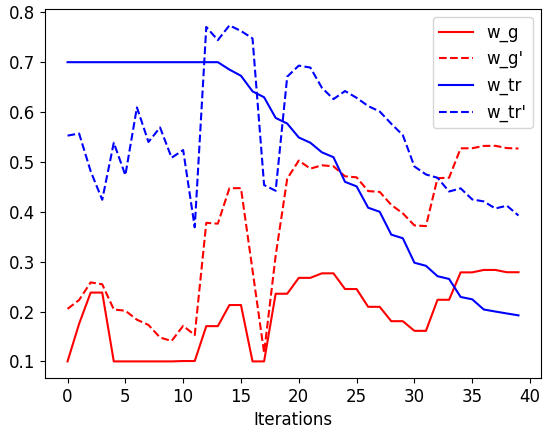}} 
    \subfloat[\label{fig:dist}]{
    \includegraphics[scale=0.3]{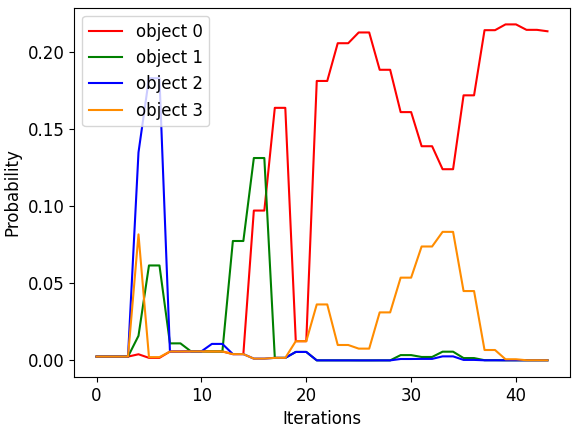}} 
    }
    \caption{(a) The set up of the shared autonomy experiment. (b) A demonstration assisted by the proposed method. (c) The weight change in Fig. b. (d) The distribution change of each goal in Fig. b.}
\label{fig:shared_autonomy}
\vspace{-0.3cm}
\end{figure}

\subsection{Experiment in the Simulation Dataset}
In this experiment, we aim to demonstrate the performance of RT on a simulation dataset. Following the practice in trajectory prediction \cite{nsp,trajectron}, we adopted the widely-used evaluation metric Average Displacement Error (ADE) and Final Displacement Error (FDE). ADE computes the average error between all the ground truth positions and the estimated positions in the trajectory, and FDE computes the displacement between the endpoints of ground truth and predicted trajectories. \emph{Best-of-20} and \emph{Most likely} trajectories are sampled to compute these metrics. To provide a point of reference, we establish a baseline using the \emph{Vanilla Residual LSTM}, which outputs the velocity of each step, with a hidden state size of 128. All the models are trained with Adam optimizer, with a learning rate of 0.001 and batch size of 256. The models are trained on a single RTX 4070Ti GPU.

\noindent \textbf{Data Collection.} We collected data with a Franka robot simulated in Pybullet.
To collect one trajectory, we randomly sample a reaching target point on the table and move the gripper to approach it along a random approach vector, from a random initial joint configuration. To control the movement, we used the MMC controller \cite{haviland2020purely}. The velocity of the gripper is calculated by:
\begin{equation}
    \bm{v}_e = k(^0\bm{T}_e^{-1} \; {^0\bm{T}}_{e*}), ~~~\bm{v}_e' = \bm{v}_e + \bm{z}*\parallel \bm{v}_e \parallel \label{noisyvelo},
\end{equation}
where $k$ is a gain term, $^0\bm{T}_e \in SE(3)$ is the end-effector instantaneous pose in the robot’s base frame, $^0\bm{T}_{e*} \in SE(3)$ is the desired end-effector pose in the robot’s base frame, $\bm{v}_e, \bm{v}_e' \in \mathbb{R}^{3}$ are the velocity and the noisy velocity of the end-effector, and $\bm{z} \in \mathbb{R}^{3}$ is a noise variable sampled from the uniform distribution $U(-1,1)$. The noisy velocity will be applied to the end-effector with a frequency of 20Hz. 
According to Eq. (\ref{noisyvelo}), the noise level is dependent on the magnitude of the velocity.
Finally, we generate 100,000 trajectories, among which 90,000 are for training and 10,000 are for testing. We name this dataset as \emph{Traj100k}.

\noindent \textbf{Performance in the Traj100k dataset.}
Results are shown in Table \ref{tab:simulation}. We only report Vanilla LSTM's most likely results since it is a deterministic method. RT performs substantially better than Vanilla LSTM in both \emph{Best-of-20} and \emph{Most likely} settings. The small error of RT demonstrates the strong capability to model the future trajectory.

\begin{table}[t]
  \centering
  \caption{The performance comparison in Traj100k.}
  \begin{tabular}{ccccc}
    \hline
    \multirow{2}*{Method}   & \multicolumn{2}{c}{Best-of-20 (mm)} & \multicolumn{2}{c}{Most likely (mm)}  \\
    \cline{2-5}
    & ADE& FDE& ADE & FDE \\
    \hline
    Vanilla LSTM \cite{lstm}    & -   & - & 136.95 & 115.47 \\
    Robot Trajectron &  17.82  & 26.97   &  30.58 & 49.94  \\
    \hline
  \end{tabular}
  \vspace{-0.2cm}
  \label{tab:simulation}
\end{table}

\noindent \textbf{Visualization of the intended point.}
Fig. \ref{fig:viz_traj} shows an example of trajectory prediction per Eq. (\ref{infer}). Both the most likely trajectory and the 2D GMMs of Eq. (\ref{infer}) are visualized. 
As the gripper moves, the generated distribution is increasingly concentrated, which indicates the increasing confidence in RT's prediction.

\begin{table}[t]
  \centering
  \caption{Shared autonomy experiment. The total time, the number of inputs and the average sum of the length of 4 trajectories in one round $l_{\text{tr}}$ are used as metrics.}
  \begin{tabular}{cccc}
    \hline
    Method & Time (sec) & Input & Average $l_{\text{tr}}$ (m)\\
    \hline
    Teleop. \cite{gottardi2022shared}    & 9.36$\pm$0.71   & 41.8$\pm$2.8 & 2.452$\pm$0.246 \\
    MaxEnt IOC \cite{gottardi2022shared} & 7.24$\pm$0.33   & 33.8$\pm$1.2 & 2.007$\pm$0.060  \\
    Robot Trajectron &  7.17$\pm$0.43  & 33.8$\pm$1.3  &  1.981$\pm$0.092  \\
    \hline
  \end{tabular}
  \vspace{-0.4cm}
  \label{tab:real}
\end{table}

\subsection{Shared Autonomy Experiment}

\noindent \textbf{Design.} In this experiment, we will compare our shared control method with two baselines. The first baseline is pure user control (named Teleop.). The other is the prevalent shared control method MaxEnt IOC \cite{javdani2018shared,muelling2017autonomy,gottardi2022shared}, for which we used the open-source code from the implementation of \cite{gottardi2022shared}. In order to make a fair comparison, we use a constant velocity to 0.1 m/s with all methods. The velocity controller MMC \cite{haviland2020purely} is leveraged to control the robot.
The experimental setting is shown in Fig. \ref{fig:setup}. 4 small cubes equipped with ArUCo markers were placed on the table. In each round, the user was required to sequentially approach the cubes on the table (4 trials) – at which point, in a real-life task, an autonomous grasp controller would take over. 
User input consists of the direction of the velocity vector, which they can control via six joystick buttons (two buttons for each axis). As written above, the velocity is kept at a constant 0.1 m/s.
Three metrics are used for comparison: the total time, the number of inputs (button pushes) and the average sum of the length of 4 trajectories in one round.

\noindent \textbf{Protocol.} We enrolled 10 novice participants from the local community. They received training in using our 3-axis joystick to control the robot. During the formal experiment, the order of control methods was randomized for each participant. When the gripper neared a goal, the robot automatically performed the grasping action.

\begin{figure}[t]
  \centering
  \includegraphics[width=\linewidth]{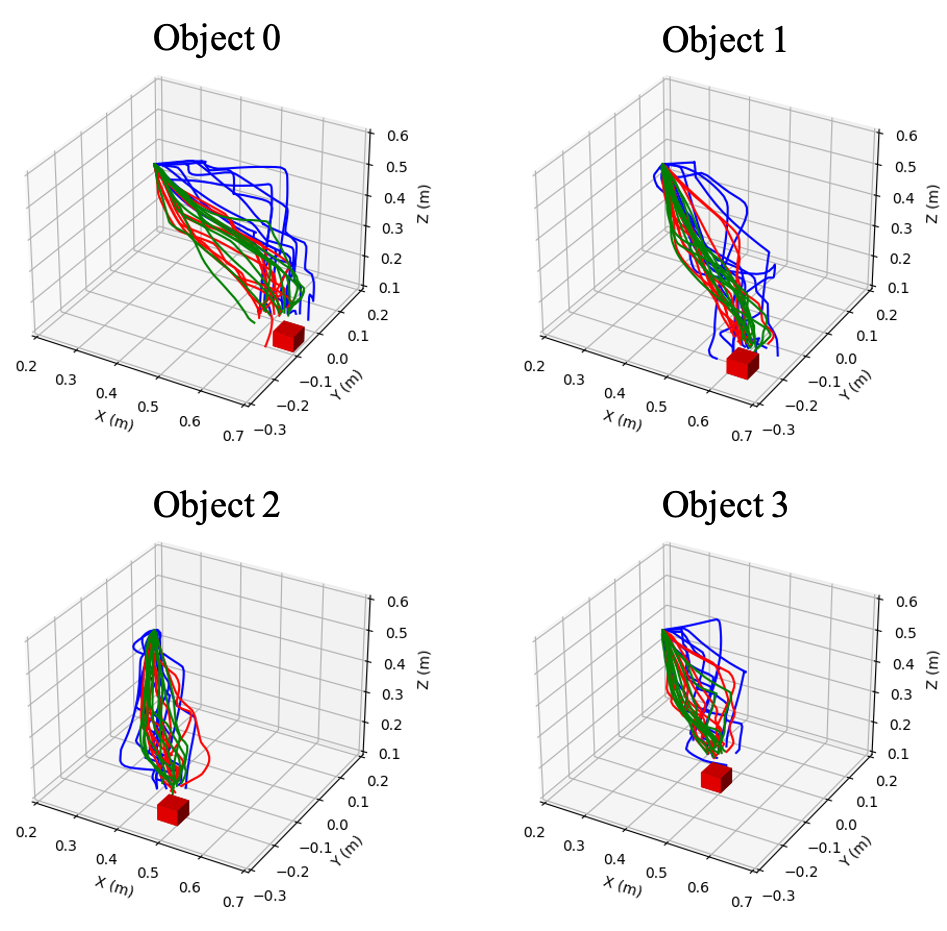} 
  \caption{User demonstrations. The blue lines denote the trajectories fully controlled by the user. The red and green lines denote the trajectories assisted by \emph{MaxEnt IOC} and \emph{Robot Trajectron}, respectively. As discussed in the text, the trajectories guided by RT to reach the object are straighter and smoother.}
  \vspace{-0.6cm}
  \label{fig:traj_cubes}
\end{figure}

\noindent \textbf{Anaylsis.} Table \ref{tab:real} shows that our proposed method uses less time and fewer inputs and produces shorter trajectories to approach the goal, demonstrating its effectiveness. 
Notably, our method achieves comparable performance to the SOTA method MaxEnt IOC. The superior performance is due to the early capturing of the robot's motion. Guided by the smooth predicted trajectory, the robot can achieve the goal faster.
Fig. \ref{fig:traj_cubes} shows the trajectories of user demonstrations for 4 cubes. Our method aids in producing smoother and more direct trajectories leading to the goal.
Fig. \ref{fig:traj} \ref{fig:weight} and \ref{fig:dist} depict the trajectory, shared weights, and goal distributions within one demonstration. Initially, the user explores the path to the goal with a low $w_g$, providing limited assistance. In this phase, It is mainly TFF that is at work. Although $w_\text{tr}$ is high, the user does not align with $v_\text{tr}$, leading to a reduction in $w_\text{tr}'$ due to the agreement mechanism. 
After the 15th iteration, Trajectron is increasingly certain about object 0, which means that GAF is at work while TFF is ceasing operation. Comparing $w_g$ and $w_g'$, we can see that the agreement mechanism then strengthens the AI's control, resulting in smoother and more efficient gripper movements towards the intended object.

\subsection{Experiment of Intent Estimation}
In this final experiment, we evaluate the model behaviors in a situation where the user changes their intent during approaching.

\begin{figure} [t!]
\centering
    \resizebox{0.9\columnwidth}{!}{
    \subfloat[\label{fig:intent_exp}]{
    \includegraphics[scale=0.4]{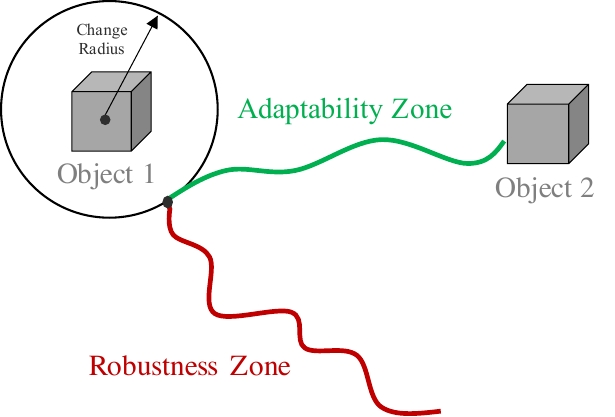}}
    \subfloat[\label{fig:intent_results}]{
    \includegraphics[scale=0.1]{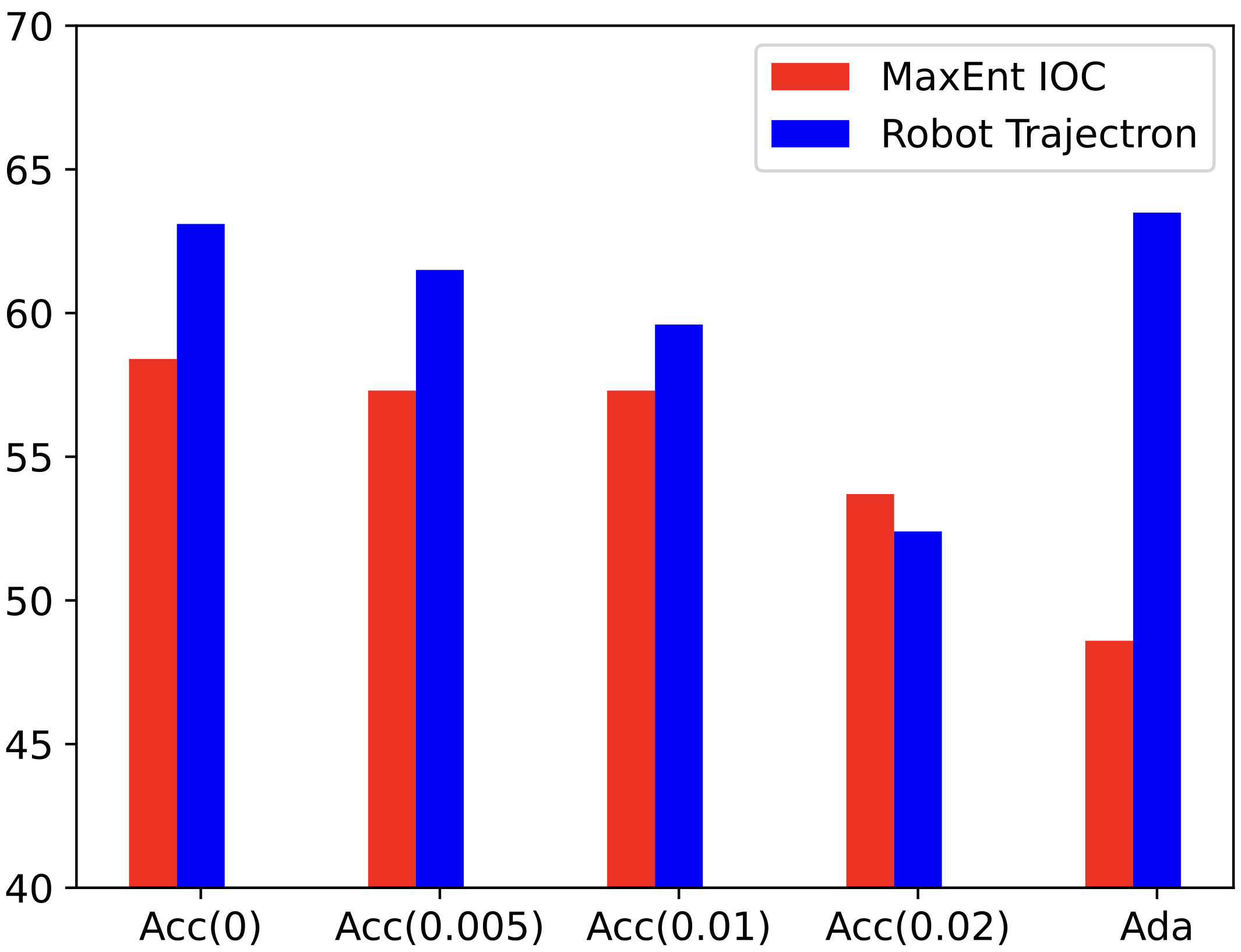}} 
    }
    \caption{(a) The illustration of the intent estimation experiment. (b) The results of the intent estimation experiment. ``Acc($\epsilon$)'' denotes the Robustness Zone accuracy in different noise levels $\epsilon$. ``Ada'' denotes the Adaptability.}
    \vspace{-0.7cm}
\label{fig:intent} 
\end{figure}

\noindent \textbf{Design.} We pre-recorded 10 change-of-intent trajectories with the same object setting as in the shared autonomy experiment. As shown in Fig. \ref{fig:intent_exp}, first, the subject will be required to approach one of the objects, and then switch to another. The first part of the motions allows us to measure the robustness of different models to user-input noise, whereas the second part allows us to measure how well a model adapts to a change of intent.
We replay these trajectories to both RT and MaxEnt IOC and evaluate their performance in the following metrics.

\noindent \textbf{Metrics.} In this experiment, we consider both adaptability and robustness as metrics for they are often conflicting qualities. 
We evaluate those as follows: (i) \textbf{Accuracy}: For each step in the pre-recorded trajectories, the intent estimation model identifies the object with the highest probability as the goal. We evaluate the accuracy of intent estimation by dividing the number of correct predictions by the total number of steps. (ii) \textbf{Robustness}: In the \emph{Robustness Zone}, we introduce noise by adding $z$ sampled from a uniform distribution $U(-\epsilon,\epsilon)$ to the trajectories. We then evaluate the accuracy of intent prediction at various noise levels $\epsilon$ as an indicator of robustness. (iii) \textbf{Adaptability}: A model with good adaptability can quickly perceive the intent change and make more accurate predictions in the \emph{Adaptability Zone}. We measure adaptability by calculating the accuracy of intent prediction in this zone. These metrics help us assess how well the models respond to intent change and their ability to maintain accurate predictions in the presence of noise.


\noindent \textbf{Analysis.} The results of the experiment are illustrated in Fig. \ref{fig:intent_results}. From the last column, the Adaptability of our method is higher than MaxEnt IOC, which is due to perceptiveness to the dynamic change. Besides, in the less noisy conditions ($\epsilon \leq 0.01m$), our method still outperforms MaxEnt IOC, because our method can capture the dynamics early on, while MaxEnt IOC still needs to move close enough to the object to make correct predictions. However, when the noise level increases to 0.02m, MaxEnt IOC outperforms our method, for the reason that RT mistakes the high noise for a signal of intent change.

\section{Conclusions}
This work addressed limitations related to the consistent-goal assumption of current shared-control works, by proposing a motion predictor that intrinsically captures user intent. By considering motion dynamics, RT can promptly adapt to changes of intent. We combine this predictor to a representation of possible goals to build a potential-field shared control solution.
We demonstrated applicability to predicting future motion trajectories, and effectiveness in shared control of a physical robot. In future work, we intend to study the applicability of our work to BCI-controlled grasping.


\bibliographystyle{IEEEtran}
\bibliography{IEEEexample}

\begin{thebibliography}{10}
\providecommand{\url}[1]{#1}
\csname url@rmstyle\endcsname
\providecommand{\newblock}{\relax}
\providecommand{\bibinfo}[2]{#2}
\providecommand\BIBentrySTDinterwordspacing{\spaceskip=0pt\relax}
\providecommand\BIBentryALTinterwordstretchfactor{4}
\providecommand\BIBentryALTinterwordspacing{\spaceskip=\fontdimen2\font plus
\BIBentryALTinterwordstretchfactor\fontdimen3\font minus \fontdimen4\font\relax}
\providecommand\BIBforeignlanguage[2]{{%
\expandafter\ifx\csname l@#1\endcsname\relax
\typeout{** WARNING: IEEEtran.bst: No hyphenation pattern has been}%
\typeout{** loaded for the language `#1'. Using the pattern for}%
\typeout{** the default language instead.}%
\else
\language=\csname l@#1\endcsname
\fi
#2}}

\bibitem{handmove}
Y.~Tamura, M.~Sugi, J.~Ota, and T.~Arai, ``Prediction of target object based on human hand movement for handing-over between human and self-moving trays,'' in \emph{ROMAN 2006-The 15th IEEE International Symposium on Robot and Human Interactive Communication}.\hskip 1em plus 0.5em minus 0.4em\relax IEEE, 2006, pp. 189--194.

\bibitem{xu2020shared}
Y.~Xu, H.~Zhang, L.~Cao, X.~Shu, and D.~Zhang, ``A shared control strategy for reach and grasp of multiple objects using robot vision and noninvasive brain--computer interface,'' \emph{IEEE Transactions on Automation Science and Engineering}, vol.~19, no.~1, pp. 360--372, 2020.

\bibitem{aarno2008motion}
D.~Aarno and D.~Kragic, ``Motion intention recognition in robot assisted applications,'' \emph{Robotics and Autonomous Systems}, vol.~56, no.~8, pp. 692--705, 2008.

\bibitem{muelling2017autonomy}
K.~Muelling, A.~Venkatraman, J.-S. Valois, J.~E. Downey, J.~Weiss, S.~Javdani, M.~Hebert, A.~B. Schwartz, J.~L. Collinger, and J.~A. Bagnell, ``Autonomy infused teleoperation with application to brain computer interface controlled manipulation,'' \emph{Autonomous Robots}, vol.~41, pp. 1401--1422, 2017.

\bibitem{gottardi2022shared}
A.~Gottardi, S.~Tortora, E.~Tosello, and E.~Menegatti, ``Shared control in robot teleoperation with improved potential fields,'' \emph{IEEE Transactions on Human-Machine Systems}, vol.~52, no.~3, pp. 410--422, 2022.

\bibitem{apf}
O.~Khatib, ``Real-time obstacle avoidance for manipulators and mobile robots,'' \emph{The international journal of robotics research}, vol.~5, no.~1, pp. 90--98, 1986.

\bibitem{nsp}
J.~Yue, D.~Manocha, and H.~Wang, ``Human trajectory prediction via neural social physics,'' in \emph{European Conference on Computer Vision}.\hskip 1em plus 0.5em minus 0.4em\relax Springer, 2022, pp. 376--394.

\bibitem{socialattn}
A.~Vemula, K.~Muelling, and J.~Oh, ``Social attention: Modeling attention in human crowds,'' in \emph{2018 IEEE international Conference on Robotics and Automation}.\hskip 1em plus 0.5em minus 0.4em\relax IEEE, 2018, pp. 4601--4607.

\bibitem{wang2007gaussian}
J.~M. Wang, D.~J. Fleet, and A.~Hertzmann, ``Gaussian process dynamical models for human motion,'' \emph{IEEE transactions on pattern analysis and machine intelligence}, vol.~30, no.~2, pp. 283--298, 2007.

\bibitem{katuwandeniya2022exact}
K.~Katuwandeniya, S.~H. Kiss, L.~Shi, and J.~V. Miro, ``Exact-likelihood user intention estimation for scene-compliant shared-control navigation,'' in \emph{2022 International Conference on Robotics and Automation (ICRA)}.\hskip 1em plus 0.5em minus 0.4em\relax IEEE, 2022, pp. 6437--6443.

\bibitem{trajectron++}
T.~Salzmann, B.~Ivanovic, P.~Chakravarty, and M.~Pavone, ``Trajectron++: Dynamically-feasible trajectory forecasting with heterogeneous data,'' in \emph{Computer Vision--ECCV 2020: 16th European Conference, Glasgow, UK, August 23--28, 2020, Proceedings, Part XVIII 16}.\hskip 1em plus 0.5em minus 0.4em\relax Springer, 2020, pp. 683--700.

\bibitem{sociallstm}
A.~Alahi, K.~Goel, V.~Ramanathan, A.~Robicquet, L.~Fei-Fei, and S.~Savarese, ``Social lstm: Human trajectory prediction in crowded spaces,'' in \emph{Proceedings of the IEEE conference on computer vision and pattern recognition}, 2016, pp. 961--971.

\bibitem{bartoli2018context}
F.~Bartoli, G.~Lisanti, L.~Ballan, and A.~Del~Bimbo, ``Context-aware trajectory prediction,'' in \emph{2018 24th international conference on pattern recognition (ICPR)}.\hskip 1em plus 0.5em minus 0.4em\relax IEEE, 2018, pp. 1941--1946.

\bibitem{socialgan}
A.~Gupta, J.~Johnson, L.~Fei-Fei, S.~Savarese, and A.~Alahi, ``Social gan: Socially acceptable trajectories with generative adversarial networks,'' in \emph{Proceedings of the IEEE conference on computer vision and pattern recognition}, 2018, pp. 2255--2264.

\bibitem{mangalam2020not}
K.~Mangalam, H.~Girase, S.~Agarwal, K.-H. Lee, E.~Adeli, J.~Malik, and A.~Gaidon, ``It is not the journey but the destination: Endpoint conditioned trajectory prediction,'' in \emph{Computer Vision--ECCV 2020: 16th European Conference, Glasgow, UK, August 23--28, 2020, Proceedings, Part II 16}.\hskip 1em plus 0.5em minus 0.4em\relax Springer, 2020, pp. 759--776.

\bibitem{trajectron}
B.~Ivanovic and M.~Pavone, ``The trajectron: Probabilistic multi-agent trajectory modeling with dynamic spatiotemporal graphs,'' in \emph{Proceedings of the IEEE/CVF International Conference on Computer Vision}, 2019, pp. 2375--2384.

\bibitem{vanhooydonck2003shared}
D.~Vanhooydonck, E.~Demeester, M.~Nuttin, and H.~Van~Brussel, ``Shared control for intelligent wheelchairs: an implicit estimation of the user intention,'' in \emph{Proceedings of the 1st international workshop on advances in service robotics (ASER’03)}, 2003, pp. 176--182.

\bibitem{goodrich2003seven}
M.~A. Goodrich and D.~R. Olsen, ``Seven principles of efficient human robot interaction,'' in \emph{SMC'03 Conference Proceedings. 2003 IEEE International Conference on Systems, Man and Cybernetics. Conference Theme-System Security and Assurance (Cat. No. 03CH37483)}, vol.~4.\hskip 1em plus 0.5em minus 0.4em\relax IEEE, 2003, pp. 3942--3948.

\bibitem{ding2011human}
H.~Ding, G.~Rei{\ss}ig, K.~Wijaya, D.~Bortot, K.~Bengler, and O.~Stursberg, ``Human arm motion modeling and long-term prediction for safe and efficient human-robot-interaction,'' in \emph{2011 IEEE International Conference on Robotics and Automation}.\hskip 1em plus 0.5em minus 0.4em\relax IEEE, 2011, pp. 5875--5880.

\bibitem{aarno2005adaptive}
D.~Aarno, S.~Ekvall, and D.~Kragic, ``Adaptive virtual fixtures for machine-assisted teleoperation tasks,'' in \emph{Proceedings of the 2005 IEEE international conference on robotics and automation}.\hskip 1em plus 0.5em minus 0.4em\relax IEEE, 2005, pp. 1139--1144.

\bibitem{schrempf2007tractable}
O.~C. Schrempf, D.~Albrecht, and U.~D. Hanebeck, ``Tractable probabilistic models for intention recognition based on expert knowledge,'' in \emph{2007 IEEE/RSJ International Conference on Intelligent Robots and Systems}.\hskip 1em plus 0.5em minus 0.4em\relax IEEE, 2007, pp. 1429--1434.

\bibitem{tahboub2006intelligent}
K.~A. Tahboub, ``Intelligent human-machine interaction based on dynamic bayesian networks probabilistic intention recognition,'' \emph{Journal of Intelligent and Robotic Systems}, vol.~45, pp. 31--52, 2006.

\bibitem{ziebart2008maximum}
B.~D. Ziebart, A.~L. Maas, J.~A. Bagnell, A.~K. Dey, \emph{et~al.}, ``Maximum entropy inverse reinforcement learning.'' in \emph{Aaai}, vol.~8.\hskip 1em plus 0.5em minus 0.4em\relax Chicago, IL, USA, 2008, pp. 1433--1438.

\bibitem{javdani2018shared}
S.~Javdani, H.~Admoni, S.~Pellegrinelli, S.~S. Srinivasa, and J.~A. Bagnell, ``Shared autonomy via hindsight optimization for teleoperation and teaming,'' \emph{The International Journal of Robotics Research}, vol.~37, no.~7, pp. 717--742, 2018.

\bibitem{cvae}
K.~Sohn, H.~Lee, and X.~Yan, ``Learning structured output representation using deep conditional generative models,'' \emph{Advances in neural information processing systems}, vol.~28, 2015.

\bibitem{beta-vae}
I.~Higgins, L.~Matthey, A.~Pal, C.~Burgess, X.~Glorot, M.~Botvinick, S.~Mohamed, and A.~Lerchner, ``beta-vae: Learning basic visual concepts with a constrained variational framework,'' in \emph{International conference on learning representations}, 2016.

\bibitem{haviland2020purely}
J.~Haviland and P.~Corke, ``A purely-reactive manipulability-maximising motion controller,'' \emph{arXiv preprint arXiv:2002.11901}, 2020.

\bibitem{lstm}
S.~Hochreiter and J.~Schmidhuber, ``Long short-term memory,'' \emph{Neural computation}, vol.~9, no.~8, pp. 1735--1780, 1997.

\end{thebibliography}

\end{document}